# 4D Multi-atlas Label Fusion using Longitudinal Images

Yuankai Huo[1], Susan M. Resnick[2] and Bennett A. Landman[1]

[1] Electrical Engineering, Vanderbilt University, Nashville, TN
[2] Laboratory of Behavioral Neuroscience, National Institute on Aging, Baltimore, MD
yuankai.huo@vanderbilt.edu

**Abstract.** Longitudinal reproducibility is an essential concern in automated medical image segmentation, yet has proven to be an elusive objective as manual brain structure tracings have shown more than 10% variability. To improve reproducibility, longitudinal segmentation (4D) approaches have been investigated to reconcile temporal variations with traditional 3D approaches. In the past decade, multi-atlas label fusion has become a state-of-the-art segmentation technique for 3D image and many efforts have been made to adapt it to a 4D longitudinal fashion. However, the previous methods were either limited by using application specified energy function (e.g., surface fusion and multi model fusion) or only considered temporal smoothness on two consecutive time points (t and t+1) under sparsity assumption. Therefore, a 4D multi-atlas label fusion theory for general label fusion purpose and simultaneously considering temporal consistency on all time points is appealing. Herein, we propose a novel longitudinal label fusion algorithm, called 4D joint label fusion (4DJLF), to incorporate the temporal consistency modeling via non-local patch-intensity covariance models. The advantages of 4DJLF include: (1) 4DJLF is under the general label fusion framework by simultaneously incorporating the spatial and temporal covariance on all longitudinal time points. (2) The proposed algorithm is a longitudinal generalization of a leading joint label fusion method (JLF) that has proven adaptable to a wide variety of applications. (3) The spatial temporal consistency of atlases is modeled in a probabilistic model inspired from both voting based and statistical fusion. The proposed approach improves the consistency of the longitudinal segmentation while retaining sensitivity compared with original JLF approach using the same set of atlases. The method is available online in open-source.

## 1    Introduction

An essential challenge in volumetric (3D) image segmentation on longitudinal medical images is to ensure the temporal consistency while retaining sensitivity. Many efforts have been made to incorporate the temporal dimension into volumetric segmentation (4D). One family of 4D methods is to control the longitudinal variations during pre/post-processing  [1]. Another family is to incorporate the longitudinal variations within segmentation methods [2]. In the past decade, multi-atlas segmentation (MAS) has been regarded as de facto standard segmentation method in 3D scenarios [3-5]. To improve the performance of 4D MAS for longitudinal data, several previous avenues have been explored[6-8]. However, these methods are restricted on surface labeling application, availability of multi-modal data, or only considering two consecutive time points ($t$ and $t$+1) while assuming the l1-norm sparsity of fusion weights. When more than two longitudinal target images are available, the more comprehensive strategy is to consider the spatial smoothness on all time points (Fig. 1).

In this paper, we propose a novel longitudinal label fusion algorithm, called 4D joint label fusion (4DJLF) to incorporate the probabilistic model of temporal performance of atlases to the voting based fusion. Briefly, we model the temporal performance of atlases on all time points in a probabilistic model and incorporate the leading and widely validated joint label fusion (JLF) framework.

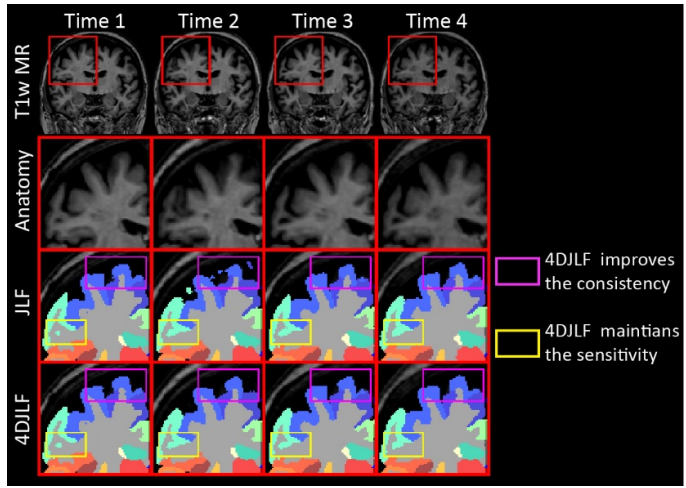

**Fig. 1.** An example of the inconsistency of 3D joint label fusion (JLF) segmentation across longitudinal multiple scans from the same subject. 4DJLF is proposed to improve the consistency while maintain the sensitivity.

## 2 Theory

### 2.1 Model Definition

A target image be represented by $T_t, t \in [1,2,...,k]$. 4DJLF considers all available longitudinal target images, $\mathbf{T} = \{T_1, T_2, ..., T_k\}$ where $T_t$ represents a target image. First, all longitudinal target images are registered to the first-time point using rigid registration [9]. $n$ pairs of atlases (one intensity atlas and one label atlas) $\mathbf{A} = \{A_1, A_2, ..., A_n\}$ are used in the MAS. Then, we register the $n$ intensity atlases to $k$ longitudinal target images to achieve $m = n \times k$ registered pairs of atlases. For mathematical convenience, we concatenate all registered atlases (based on the sequence in $\mathbf{T}$) to derive $m$ registered intensity atlases set $\mathbf{I}$ and $m$ registered label atlases set $\mathbf{S}$ as

$$\mathbf{I} = \{I_1^{(1)}, ..., I_n^{(1)}, I_{n+1}^{(2)}, ..., I_{2n}^{(2)}, \cdots, I_{2n+1}^{(k)}, ..., I_m^{(k)}\}$$

$$\mathbf{S} = \{S_1^{(1)}, ..., S_n^{(1)}, S_{n+1}^{(2)}, ..., S_{2n}^{(2)}, \cdots, S_{2n+1}^{(k)}, ..., S_m^{(k)}\} \tag{1}$$

where the superscripts "$(\cdot)$" indicate to which target image that atlas was registered.

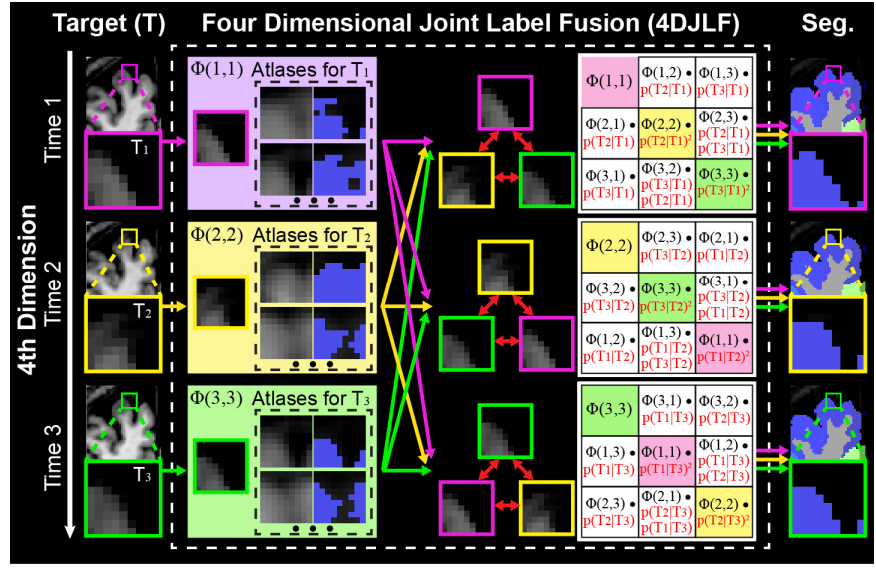

**Fig. 2.** The 4DJLF framework. First, the same set of atlases are registered to the longitudinal target images (3 time points in figure). Then, the $\Phi$ matrices are calculated using Eq. (9). Finally, the spatial temporal performance of all atlases are model by Eq. (10), which leads to the final segmentations ("Seg."). Note that the upper right $3 \times 3$ matrix is identical to Eq. (11). The original JLF estimates the block diagonal elements of the generalized covariance matrix (highlighted in magenta, green, and yellow) which would result in independent temporal estimates.

The $k$ longitudinal target images provide $m$ registered atlases, where each atlas corresponds to one time point (target image). The consensus segmentation $\bar{S}$ for voxel $x$ on $t_{th}$ target image is $\bar{S}^t(x) = \sum_{i=1}^m w_i^t(x) S_i(x) = \mathbf{w}^t(x) \cdot \mathbf{S}(x)$, where $\mathbf{w}^t(x) = \{w_1^t(x), w_2^t(x), ..., w_m^t(x)\}$ are spatially varying weights restricted by $\sum_{i=1}^m w_i^k(x) = 1$. Adopting [10], the error $\delta_i^t(x)$ made by atlas $S_i$ on $t_{th}$ target image in the binary segmentation is $\delta_i^t(x) = S_T^t(x) - S_i(x)$, where $S_T^t(x)$ is the hidden true segmentation. $\delta_i^t(x) = 0$ indicates the right decision is made, while $\delta_i^t(x) = -1$ or $1$ means the wrong decision is made. Then, our purpose is to find a set of voting weights $\mathbf{w}^t(x)$ for each target image $T_t$ that minimize the total expected error between the automated labeled image $\bar{S}^k$ and hidden true $S_T^t$, given by the following energy function

$$E_{\delta_1^t(x),...,\delta_m^t(x)} \left[ \left( S_T^t(x) - \bar{S}^t(x) \right)^2 | \mathbf{T}, \mathbf{I} \right] = \tag{2}$$

$$= \sum_{i=1}^m \sum_{j=1}^m w_i^t(x) w_j^t(x) E_{\delta_i^t(x)\delta_j^t(x)} [\delta_i^t(x) \delta_j^t(x) | T_1, ..., T_k, I_1, ..., I_m] = \mathbf{w}_x^{t\,T} \mathbf{M}_x^t \mathbf{w}_x^t$$

where $\mathbf{w}_x^{t\,T}$ is the transpose of vector $\mathbf{w}_x^t$ at voxel $x$. $\boldsymbol{M}_x^t$ is a $m \times m$ pairwise dependency matrix that $\boldsymbol{M}_x^t(i,j) = p(\delta_i^t(x)\delta_j^t(x) = 1|T_1, \dots, T_k, l_1, \dots, l_m)$. Finally, the estimated weights $\widehat{\mathbf{w}}_x^t$, which is our target, is derived by $\widehat{\mathbf{w}}_x^t = \arg\min_{\mathbf{w}_x^t} \mathbf{w}_x^{t\,T}(\boldsymbol{M}_x^t + \alpha \boldsymbol{I})\mathbf{w}_x^t$

## 2.2    JLF-Multi

As a baseline, we consider to use simple temporal model (JLF-Multi) to perform the 4D label fusion. We assume that each target image in $\mathbf{T}$ contributes equally to the label fusion for target $T_t$. In this case, $\boldsymbol{M}_x^t(i,j)$ is can be approximated as

$$\boldsymbol{M}_x^t(i,j) \propto \sum_{y \in B(x)} |T_t(y) - I_i(\mathcal{N}_i(y))| \cdot |T_t(y) - I_j(\mathcal{N}_j(y))| \tag{3}$$

where the $\Sigma$ improves the spatial smoothness by adding multiple voxels $y$ in a patch neighborhood $B(x)$ (e.g., $2 \times 2 \times 2$ by default), and the non-local patch searching is conducted within a search neighborhood $\mathcal{N}(y)$ (e.g., $3 \times 3 \times 3$ by default).

## 2.3    4DJLF

In JLF-Multi, each longitudinal target image contributes equally to the 4D label fusion. However, this assumption is not always valid. Herein, we propose the new dependency matrix $\check{\boldsymbol{M}}_x^t(i,j)$ by adaptively evaluating the longitudinal raters' performance on any target image patches using a probabilistic model

$$\check{\boldsymbol{M}}_x^t(i,j) = \tag{4}$$

$$p\left(T_q(x), T_r(x)\big|T_t(x)\right) \cdot \left(\sum_{y \in B(x)} \left|T_q(y) - I_i^{(q)}\big(\mathcal{N}_i(y)\big)\right| \cdot \left|T_r(y) - I_j^{(r)}\big(\mathcal{N}_j(y)\big)\right|\right)$$

where the new dependency matrix $\check{\boldsymbol{M}}_x^t(i,j)$ not only evaluates the similarity between atlases and target images but also considers the longitudinal similarities between target images. The "$(q)$" and "$(r)$" indicate which atlases that $I_i$ and $I_j$ were registered to and the value of $q$ and $r$ are derived from Eq. (1). Then, probability of using the raters (atlases) from $T_q$ and $T_r$ given target $T_t$ is modeled in a conditional probability

$$p\left(T_q(x), T_r(x)\big|T_t(x)\right) = p(T_q(x)|T_t(x)) \cdot p(T_r(x)|T_t(x)) \tag{5}$$

by assuming $T_q$ and $T_r$ are conditionally independent, we have

$$p\left((T_q(x)|T_t(x))\right) = \exp\left(\beta \cdot \sum_{y \in B(x)} \frac{|T_q(y) - T_t(y)|}{\left|T_q(y) - I_i^{(q)}(\mathcal{N}_i(y))\right|}\right) \tag{6}$$

$$p\left((T_r(x)|T_t(x))\right) = \exp\left(\beta \cdot \sum_{y \in B(x)} \frac{|T_r(y) - T_t(y)|}{\left|T_r(y) - I_j^{(r)}(\mathcal{N}_j(y))\right|}\right) \tag{7}$$

where $\beta$ is a sensitivity coefficient and is empirically set to 100 in the experiments.

### 2.4 Relationship between 4DJLF to JLF

The proposed 4DJLF theory is a generalization of JLF. If the $\beta$ is set to a large number, the $p\left(T_q(x), T_r(x)\big|T_t(x)\right)$ will be large for atlases from other time points, but still equals to 1 for the atlases from the target image itself. Therefore, the weights of the atlases from other time points will be close to zero and essentially only the atlases registered to the target time $T_t$ are considered. In that case, 4DJLF degenerates to JLF. To see the relationship in Fig. 2, we redefine the right side of Eq. (5).

$$\Gamma_x(i,j) = \sum_{y \in B(x)} \cdot \left|T_q(y) - I_i^{(q)}\left(\mathcal{N}_i(y)\right)\right| \cdot \left|T_r(y) - I_j^{(r)}\left(\mathcal{N}_j(y)\right)\right| \tag{8}$$

Then, we define a matrix $\Phi_{p,q}$ as the following

$$\Phi_x(q,r) = \begin{bmatrix} \Gamma_x(i',j') & \Gamma_x(i',j'+1) & \cdots & \Gamma_x(i',j'+k) \\ \Gamma_x(i'+1,j') & \Gamma_x(i'+1,j'+1) & & \Gamma_x(i'+1,j'+k) \\ \vdots & & \ddots & \vdots \\ \Gamma_x(i'+k,j') & \Gamma_x(i'+k,j'+1) & \cdots & \Gamma_x(i'+k,j'+k) \end{bmatrix} \tag{9}$$

where $i' = (q-1) \times k + 1$ and $j' = (r-1) \times k + 1$. For simplify, we assume three longitudinal target images are used and the first time point is the target image (upper row in Fig. 2). We rewrite the $p\left(\left(T_q(x)\big|T_t(x)\right)\right)$ as $p_x\left(\frac{T_q}{T_t}\right)$ to visualize the $\check{M}^t$ at the first time point ($t = 1$ and the subscript $x$ is omitted for simplicity). Since $p\left(\frac{T_1}{T_1}\right) = 1$, the $\check{M}^1$ is further simplified to

$$\check{M}^1 = \begin{bmatrix} \Phi(1,1) & \Phi(1,2)p\left(\frac{T_2}{T_1}\right) & \Phi(1,3)p\left(\frac{T_3}{T_1}\right) \\ \Phi(2,1)p\left(\frac{T_2}{T_1}\right) & \Phi(2,2)p\left(\frac{T_2}{T_1}\right)^2 & \Phi(2,3)p\left(\frac{T_2}{T_1}\right)p\left(\frac{T_3}{T_1}\right) \\ \Phi(3,1)p\left(\frac{T_3}{T_1}\right) & \Phi(3,2)p\left(\frac{T_3}{T_1}\right)p\left(\frac{T_2}{T_1}\right) & \Phi(3,3)p\left(\frac{T_3}{T_1}\right)^2 \end{bmatrix} \tag{10}$$

where $\check{M}^1$ is identical to the upper right matrix in Fig. 2. Note that $\Phi(1,1)$ is the same as the $M_x$ in JLF [10], which demonstrates the relationship between 4DJLF and JLF.

## 3 Experimental Methods and Results

Six healthy subjects with 21 longitudinal T1-weighted (T1w) MR scans (mean age 82.3, range:72.5~90.2) were randomly selected from Baltimore Longitudinal Study of Aging (BLSA) [11]. Each image had 170×256×256 voxels with 1.2×1×1 mm resolution. 15 pairs of atlases from BrainCOLOR (http://braincolor.mindboggle.info/protocols/) were employed. The intensity atlases had 1mm isotropic resolution and the label atlases contained 132 labels. In order to evaluate the sensitivity, one randomly selected T1w image from a healthy subject (age 11) in ADHD-200 OHSU dataset (http://fcon_1000.projects.nitrc.org/indi/adhd200/) was used in the robustness test. The 21 longitudinal target images were first affinely registered to the MNI305 atlas. Then, the spatially aligned longitudinal atlases $\mathbf{T} = \{T_1, T_2, ..., T_k\}$ were derived by rigidly registering each target image to the first time point. Then, 15 atlases were non-rigidly registered [12] to all target images to achieve the intensity and label atlases in Eq. (1) (performed $m = 15 \times 21$ non-rigid registrations). The same preprocessing was also deployed to the one ADHD-200 target image.

JLF was deployed on all 21 longitudinal target images independently using default parameters. The longitudinal reproducibility of JLF, JLF-multi and 4D JLF were evaluated by calculating the Dice similarity coefficients between all pairs of longitudinal images (Fig. 3a) Wilcoxon signed rank test and Cohen's d effect size were performed on JLF-Multi vs. JLF and 4D vs. JLF. The "*" indicated the difference satisfied (1) p<0.01 in Wilcoxon signed rank test, and (2) d>0.1 in effect size. The temporal changes on volume sizes of whole brain, gray matter and white matter were shown in Fig. 4. Fig. 5 shown the qualitive results from subject 5 in Fig. 4.

Second, a robustness test was conducted to evaluate the sensitivity of JLF, JLF-Multi and 4DJLF. We combined the previously mentioned ADHD-200 image to each target image to formed 21 dummy longitudinal pairs. This test simulated the large temporal variations since the two images in each pair were independent and collected from different scanners. Then, the 4D segmentation methods were deployed on such cases to see if the 4D methods can maintain the sensitivity compared with JLF. The Fig. 3b indicated the 4DJLF had "trivial" changes on reproducibility (effect size <0.1 compared with JLF, while JLF-Multi had large differences compared with JLF.

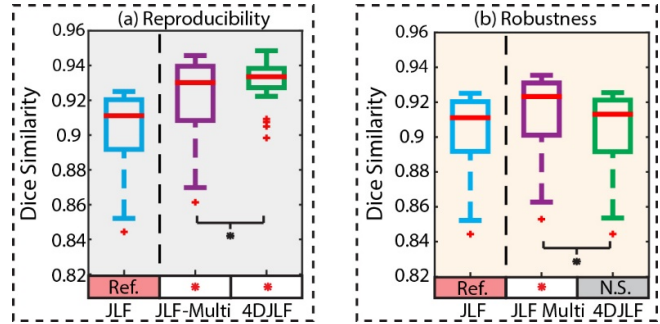

**Fig. 3.** Quantitative results. (a) The reproducibility experiment shown that the proposed 4DJLF had overall significantly better reproducibility than JLF and JLF-Multi. (b) The robustness test indicated that 4DJLF maintained the sensitivity as JLF, while JLF-Multi was not able to do so. The red "*" means the method satisfied both p<0.01 and effect size>0.1 compared with JLF ("Ref."), while the "N.S." means at least one was not satisfied. The black "*" means the difference between two methods satisfied both p<0.01 and effect size>0.1.

## 4 Conclusion

We propose the 4DJLF multi-atlas label fusion strategy by modeling the spatial temporal performance of atlases. The proposed theory incorporates the ideas from the two major families of label fusion theories (voting based fusion and statistical fusion) by generalizing the JLF label fusion method to a 4D manner. The results demonstrated that the proposed method was not only able to improve the longitudinal reproducibility (Fig. 3a, 4 and 5) but also reduces the segmentation errors compared with traditional 3D JLF (Fig. 5). Meanwhile, the 4DJLF did not significantly change the segmentation reproducibility when performing on dummy longitudinal pairs of images (Fig 3b).

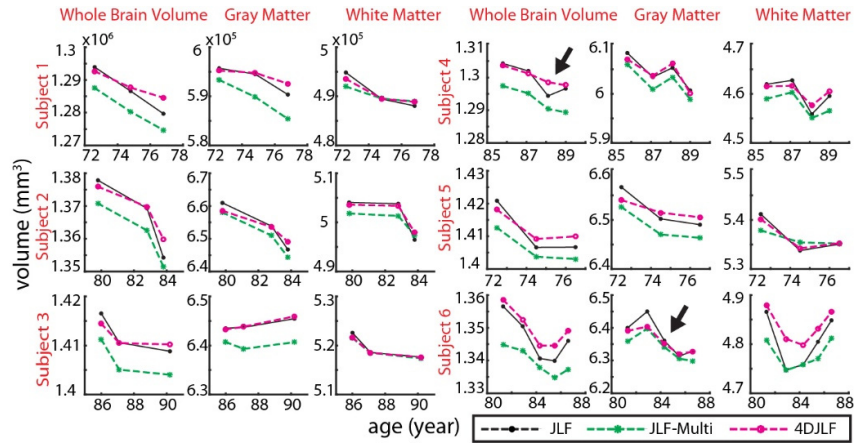

**Fig. 4.** This figure presents the longitudinal changes of whole brain volume, gray matter volume and white matter volume for all 6 subjects (21 time points). The black arrows indicate that the proposed 4DJLF reconciles some obvious temporal inconsistency by simultaneously considering all available longitudinal images.

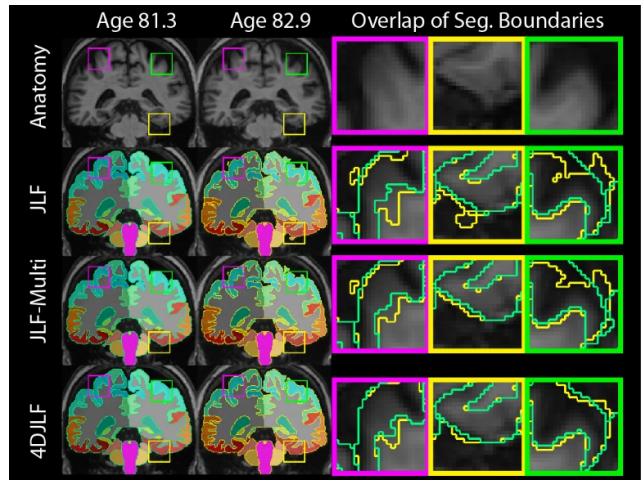

**Fig. 5.** Qualitative results of deploying longitudinal segmentation methods on two examples.

**Acknowledgments**. This research was supported by NSF CAREER 1452485, NIH 5R21EY024036, NIH 1R21NS064534, NIH 2R01EB006136, NIH 1R03EB012461, and supported by the Intramural Research Program, National Institute on Aging, NIH. This project was supported in part by the National Center for Research Resources, Grant UL1 RR024975-01, and is now at the National Center for Advancing Translational Sciences, Grant 2 UL1 TR000445-06. The content is solely the responsibility of the authors and does not necessarily represent the official views of the NIH.